\documentclass{article}
\pdfoutput=1
\PassOptionsToPackage{numbers, compress}{natbib}

\usepackage[final]{neurips_2019}

\usepackage[utf8]{inputenc} 
\usepackage[T1]{fontenc}    
\usepackage{hyperref}       
\usepackage{url}            
\usepackage{booktabs}       
\usepackage{amsfonts}       
\usepackage{nicefrac}       
\usepackage{microtype}      
\usepackage{graphicx}
\usepackage{caption}
\usepackage{subfigure}
\usepackage{url}
\usepackage{amsmath}
\graphicspath{ {plots/} }
\usepackage{xcolor}
\usepackage{authblk}
\usepackage{appendix}

\title{A Neural Topic-Attention Model for \\Medical Term Abbreviation Disambiguation}

\author[1]{\textbf{Irene Li}}
\author[2]{\textbf{Michihiro Yasunaga\thanks{Work done at Yale University.}~ }}
\author[1]{\textbf{Muhammed Yavuz Nuzumlalı}}
\author[3]{\textbf{Cesar Caraballo}}
\author[3]{{\textbf{Shiwani\,\,Mahajan}}}
\author[3]{\textbf{Harlan Krumholz}}
\author[1]{\textbf{Dragomir Radev}}
\affil[1]{Department of Computer Science, Yale University}
\affil[2]{Department of Computer Science, Stanford University}
\affil[3]{School of Medicine,
  Yale University}
\affil[1]{\texttt {\{irene.li,yavuz.nuzumlali,dragomir.radev\}@yale.edu}}
\affil[2]{\texttt{myasu@cs.stanford.edu}}
\affil[3]{\texttt {\{cesar.caraballo,shiwani.mahajan,harlan.krumholz\}@yale.edu}}

\begin{document}
\maketitle

\begin{abstract}
  Automated analysis of clinical notes is attracting increasing attention. However, there has not been much work on medical term abbreviation disambiguation. Such abbreviations are abundant, and highly ambiguous, in clinical documents. One of the main obstacles is the lack of large scale, balance labeled data sets. To address the issue, we propose a few-shot learning approach to take advantage of limited labeled data. Specifically, a neural topic-attention model is applied to learn improved contextualized sentence representations for medical term abbreviation disambiguation. Another vital issue is that the existing scarce annotations are noisy and missing.  We re-examine and correct an existing dataset for training and collect a test set to evaluate the models fairly especially for rare senses.  We train our model on the training set which contains 30 abbreviation terms as categories (on average, 479 samples and 3.24 classes in each term) selected from a public abbreviation disambiguation dataset, and then test on a manually-created balanced dataset (each class in each term has 15 samples). We show that enhancing the sentence representation with topic information improves the performance on small-scale unbalanced training datasets by a large margin, compared to a number of baseline models. 
  
\end{abstract}

\section{Introduction}

Medical text mining is an exciting area and is becoming attractive to natural language processing (NLP) researchers. Clinical notes are an example of text in the medical area that recent work has focused on \cite{hughes2017medical,weng2017medical,yao2019clinical}. This work studies abbreviation disambiguation on clinical notes \cite{liu2002automatic,liu2001disambiguating},
specifically those used commonly by physicians and nurses.
Such clinical abbreviations can have a large number of meanings, depending on the specialty \cite{xu2015clinical,joopudi2018convolutional}. For example, the term \textit{MR} can mean \textit{magnetic resonance}, \textit{mitral regurgitation}, \textit{mental retardation}, \textit{medical record} and the general English \textit{Mister (Mr.)}. Table \ref{example} illustrates such an example. 
Abbreviation disambiguation is 
an important task in medical text understanding task \cite{jiang2011study}. Successful recognition of the abbreviations in the notes can contribute to downstream tasks such as classification, named entity recognition, and relation extraction \cite{jiang2011study}. 

\begin{table}[h]
\centering
\small
\begin{tabular}{|l|}\hline
\textbf{Note 1} \\
...72 year-old male with history of DM2 (Diabetes  
Mellitus Type 2), myocardial infarction  requiring \\
CABG(coronary artery bypass graft), asthma, \textcolor{red}{MR},  
and germ cell tumor with metastases to left upper lobe... \\
\hline
\hline
\textbf{Note 2} \\...She also underwent an echocardiogram which showed left ventricular systolic function which was normal. \\
She had mild \textcolor{red}{MR} and mild TR. She had some early diastolic dysfunction as well as biatrial enlargement...\\
\hline
\end{tabular}
\caption{\small{Examples showing the abbreviation term \textit{MR}: \textit{MR} from Note 1 is the abbreviation of \lq mental retardation\rq. One could easily misinterpret it as \lq mitral regurgitation\rq which may lead to further imaging studies with subsequent delaying or avoidance of non-urgent interventions due to higher risk of complications. In Note 2, \textit{MR} means \lq mitral regurgitation\rq \cite{xu2015clinical}.} \vspace{-8mm} }
\label{example}
\end{table}

Recent work focuses on formulating the abbreviation disambiguation task as a classification problem, where the possible senses of a given abbreviation term are pre-defined with the help of domain experts \cite{joopudi2018convolutional,xu2015clinical}. Traditional features such as part-of-speech (POS) and Term Frequency-Inverse Document Frequency (TF-IDF) are widely investigated for clinical notes classification. Classifiers like support vector machines (SVMs) and random forests (RFs) are used to make predictions \cite{weng2017medical}. Such methods depend heavily on feature engineering. Recently, deep features have been investigated in the medical domain. Word embeddings \cite{mikolov2013distributed} and Convolutional Neural Networks (CNNs) \cite{kim2014convolutional,krizhevsky2012imagenet} provide very competitive performance on text classification for clinical notes and abbreviation disambiguation task \cite{hughes2017medical,liu2018exploiting,baumel2017multi,joopudi2018convolutional}. Beyond vanilla embeddings, \cite{jin2019deep} utilized contextual features to do  abbreviation expansion. Another challenge is the difficulty in obtaining training data: clinical notes have many restrictions due to privacy issues and require domain experts to develop high-quality annotations,  thus leading to limited annotated training and testing data.  
Another difficulty is that in the real world (and in the existing public datasets), some abbreviation term-sense pairs (for example, \textit{AB} as \textit{abortion}) have a very high frequency of occurrence \cite{xu2015clinical}, while others are rarely found. This long tail issue creates the challenge of training under unbalanced sample distributions. We tackle these problems in the setting of few-shot learning \cite{garcia2017few,khodadadeh2018unsupervised} where only a few or a low number of samples can be found in the training dataset to make use of limited resources. We propose a model that combines deep contextual features based on ELMo \cite{Peters:2018} and topic information to solve the abbreviation disambiguation task.


Our contributions can be summarized as: 1) we re-examined and corrected an existing dataset for training and we collected a test set for evaluation with focus especially for rare senses; 2) we proposed a few-shot learning approach which combines topic information and contextualized word embeddings to solve clinical abbreviation disambiguation task. The implementation is available online\footnote{\url{https://github.com/IreneZihuiLi/TopicAttentionMedicalAD.git}}; 3) as limited research are conducted on this particular task, we  evaluated and compared a number of baseline methods including classical models and deep models comprehensively.

\section{Datasets}
\vspace{-3mm} 

\textbf{Training Dataset} UM Inventory \cite{xu2015clinical} is a public dataset created by researchers from the University of Minnesota, containing about 37,500 training samples with 75 abbreviation terms. Existing work reports abbreviation disambiguation results on 50 abbreviation terms \cite{joopudi2018convolutional,xu2015clinical,moon2012automated}. However, after carefully reviewing this dataset, we found that it contains many samples where medical professionals disagree: wrong samples and uncategorized samples\footnote{Wrong cases include obvious wrong abbreviation labels. For instance, under the term \textit{LE}, we found samples where the true term label should be \textit{LV}. Uncategorized samples are some sentences which have no labels.}. Due to these mistakes and flaws of this dataset, we removed the erroneous samples and eventually selected 30 abbreviation terms as our training dataset that can be made public. Among all the abbreviation terms, we have 11,466 samples, and 93 term-sense pairs in total (on average 123.3 samples/term-sense pair and 382.2 samples/term). Some term-sense pairs are very popular with a larger number of training samples but some are not (typically less than 5), we call them \textit{rare-sense} cases \footnote{Rare-senses like \textit{AC} as \textit{acetate} (2 samples in the training dataset), \textit{PE} as \textit{pleural effusion} (only 1 sample can be found in the training dataset).}. More details can be found in Appendix \ref{appendix:graph}.  


\textbf{Testing Dataset} Our testing dataset takes MIMIC-III \cite{johnson2016mimic} and PubMed\footnote{\url{https://www.ncbi.nlm.nih.gov/pubmed/}} as data sources. Here we are referring to all the notes data from MIMIC-III (NOTEEVENTS table \footnote{\url{https://mimic.physionet.org/mimictables/noteevents/}}) and \textit{Case Reports} articles from PubMed, as this type of contents are close to medical notes. We provide detailed information in Appendix \ref{app:testing}, including the steps to create the testing dataset. Eventually, we have a balanced testing dataset, where each term-sense pair has at least 11 and up to 15 samples for training (on average, each pair has 14.56 samples and the median sample number is 15).

\section{Baselines} \label{bas}

We conducted a comprehensive comparison with the baseline models, and some of them were never investigated for the abbreviation disambiguation task. 
We applied traditional features by simply taking the TF-IDF features as the inputs into the classic classifiers. Deep features are also considered: a Doc2vec model \cite{le2014distributed} was pre-trained using Gensim\footnote{\url{https://radimrehurek.com/gensim/models/doc2vec.html}. The dimension is 100.} and these word embeddings were applied to initialize deep models and fine-tuned. 

\textbf{TF-IDF}: We applied TF-IDF features as inputs to four classifiers: support vector machine classifier (\textit{SVM}), logistic regression classifier (\textit{LR}), Naive Bayes classifier (\textit{NB}) and random forest (\textit{RF}); \textbf{CNN}: We followed the same architecture as \cite{kim2014convolutional}; \textbf{LSTM}: We applied an LSTM model \cite{hochreiter1997long} to classify the sentences with pre-trained word embeddings;\textbf{LSTM-soft}: We then added a soft-attention \cite{luong2015effective} layer on top of the \textit{LSTM} model where we computed soft alignment scores over each of the hidden states; \textbf{LSTM-self}: We applied a self-attention layer \cite{bahdanau2014neural} to \textit{LSTM} model. We denote the content vector as $c_i$ for each sentence $i$, as the input to the last classification layer.



\section{Topic-attention Model}
\vspace{-3mm}


 

\textbf{ELMo} ELMo is a new type of word representation introduced by \cite{Peters:2018} which considers contextual information captured by pre-trained BiLSTM language models. Similar works like BERT \cite{devlin2018bert} and BioBERT \cite{lee2019biobert} also consider context but with deeper structures. Compared with those models, ELMo has less parameters and not easy to be overfitting. 
We trained our ELMo model on the MIMIC-III corpus. 
Since some sentences also appear in the test set, one may raise the concern of performance inflation in testing.  However, we pre-trained ELMo using the whole corpus of MIMIC in an unsupervised way, so the classification labels were not visible during training. To train the ELMo model, we adapted code from AllenNLP\footnote{\url{https://github.com/allenai/allennlp/blob/master/tutorials/how_to/elmo.md}}, we set the word embedding dimension to 64 and applied two BiLSTM layers. For all of our experiments that involved with ELMo, we initialized the word embeddings from the last layer of ELMo.

\textbf{Topic-attention} We propose a neural topic-attention model for text classification. Our assumption is that the topic information plays an important role in doing the disambiguation given a sentence. For example, the abbreviation of the medical term \textit{FISH} has two potential senses: general English \textit{fish} as seafood and the sense of \textit{fluorescent in situ hybridization} \cite{moon2012automated}. The former case always goes with the topic of food, allergies while the other one appears in the topic of some examination test reports. 
In our model, a topic-attention module is applied to add topic information into the final sentence representation.
This module calculates the distribution of topic-attention weights on a list of topic vectors. As shown in Figure \ref{model}, we took the content vector $c_i$ (from soft-attention \cite{bahdanau2014neural}) and a topic matrix $T_{topic}=[t_1,t_2,..,t_j]$ (where each $t_i$ is a column vector in the figure and we illustrate four topics) as the inputs to the topic-attention module, and then calculated a weighted summation of the topic vectors. The final sentence representation $r_i$ for the sentence $i$ was calculated as the following:
\begin{equation}
    \begin{aligned} v _ { i j } = \tanh \left( c_{i}^T W _ { topic } t _ { j } + b _ { topic } \right), ~~
    \beta _ { i j } = \frac { \exp \left( v _ { i j } ^ { \top } v _ { w } \right) } { \sum _ { j } \exp \left( v _ { i j } ^ { \top } v _ { w } \right) }, ~~ 
    s _ { i }  = \sum _ { j } \beta _ { i j } t _ { j }, ~~ 
    r_i = [c_i,s_i]
    \end{aligned}
    \label{topic}
\end{equation}
where $W_{topic}$ and $b _ { topic }$ are trainable parameters, $\beta_{it}$ represents the topic-attention weights. The final sentence representation is denoted as $r_i$, and $[\cdot,\cdot]$ means concatenation.
Here $s_i$ is the representation of the sentence which includes the topic information. The final sentence representation $r_i$ is the concatenation of $c_i$ and $s_i$, now we have both context-aware sentence representation and topic-related representation. Then we added a fully-connected layer, followed by a Softmax to do classification with cross-entropy loss function. 

\textbf{Topic Matrix}
To generate the topic matrix $T_{topic}$ as in Equation \ref{topic}, we propose a convolution-based method to generate topic vectors. We first pre-trained a topic model using the Latent Dirichlet Allocation (LDA) \cite{blei2003latent} model on MIMIC-III notes as used by the ELMo model. We set the number of topics to be 50 and ran for 30 iterations. Then we were able to get a list of top $k$ words (we set $k=100$) for each topic. 
To get the topic vector $t$ for a specific topic:
\begin{equation}
\vspace{-2mm}
    \begin{aligned} 
    t = \mathrm{maxpool}(\mathrm{Conv}(\mathbf{E})), ~~ \mathbf{E}=[e_1,e_2,..e_k]
    \end{aligned}
    \label{topic_matrix}
\end{equation}

where $e_j$  (column vector) is the pre-trained Doc2vec word embedding for the top word $j$ from the current topic, and $Conv(\cdot)$ indicates a convolutional layer; $maxpool(\cdot)$ is a max pooling layer. We finally reshaped the output $t$ into a 1-dimension vector. Eventually we collected all topic vectors as the topic matrix $T_{topic}$ in Figure \ref{model}.



\vspace{-3mm}

\section{Evaluation}
\vspace{-2mm}
We did three groups of experiments including two baseline groups and our proposed model. The first group was the TF-IDF features in Section \ref{bas} for \textit{traditional models}. The Na\"ive Bayesian classifier (\textit{NB}) has the highest scores among all traditional methods. The second group of experiments used neural models described in Section \ref{bas}, where LSTM with self attention model (\textit{LSTM-self}) has competitive results among this group, we choose this model as our base model. 
Notably, this is the first study that compares and evaluates LSTM-based models on the medical term abbreviation disambiguation task.

\begin{minipage}[b]{.45\textwidth}
  \centering
  \small
  \begin{tabular}{|lcc|}\hline
\textbf{Method}       & \textbf{Accuracy} & \textbf{F1(macro)}     \\\hline
\textit{Traditional Models} & &\\
SVM          & 0.3700   & 0.2103 \\
LR           & 0.4852   & 0.3689 \\
RF           & 0.5303   & 0.4469 \\
NB           & \textbf{0.5801}   & \textbf{0.5225} \\ \hline

\textit{Neural Models} & & \\ 
CNN          & 0.5838   & 0.5169 \\
LSTM         & 0.6154   & 0.5520 \\
LSTM-soft         & 0.5096   & 0.4328 \\
LSTM-self    & \textbf{0.6249}   & \textbf{0.5555} \\\hline
\textit{Proposed Model} & & \\ 
Topic Only    & 0.6985   & 0.6524 \\
ELMo Only     & 0.7236   & 0.6781 \\
ELMo+Topic   & \textbf{0.7476}   & \textbf{0.7041} \\\hline
\end{tabular}
\captionof{table}{Experimental Results: we report macro F1 in all the experiments.}
\label{tab:results}
\end{minipage}\qquad
~
\begin{minipage}[b]{.45\textwidth}
\includegraphics[width=6cm]{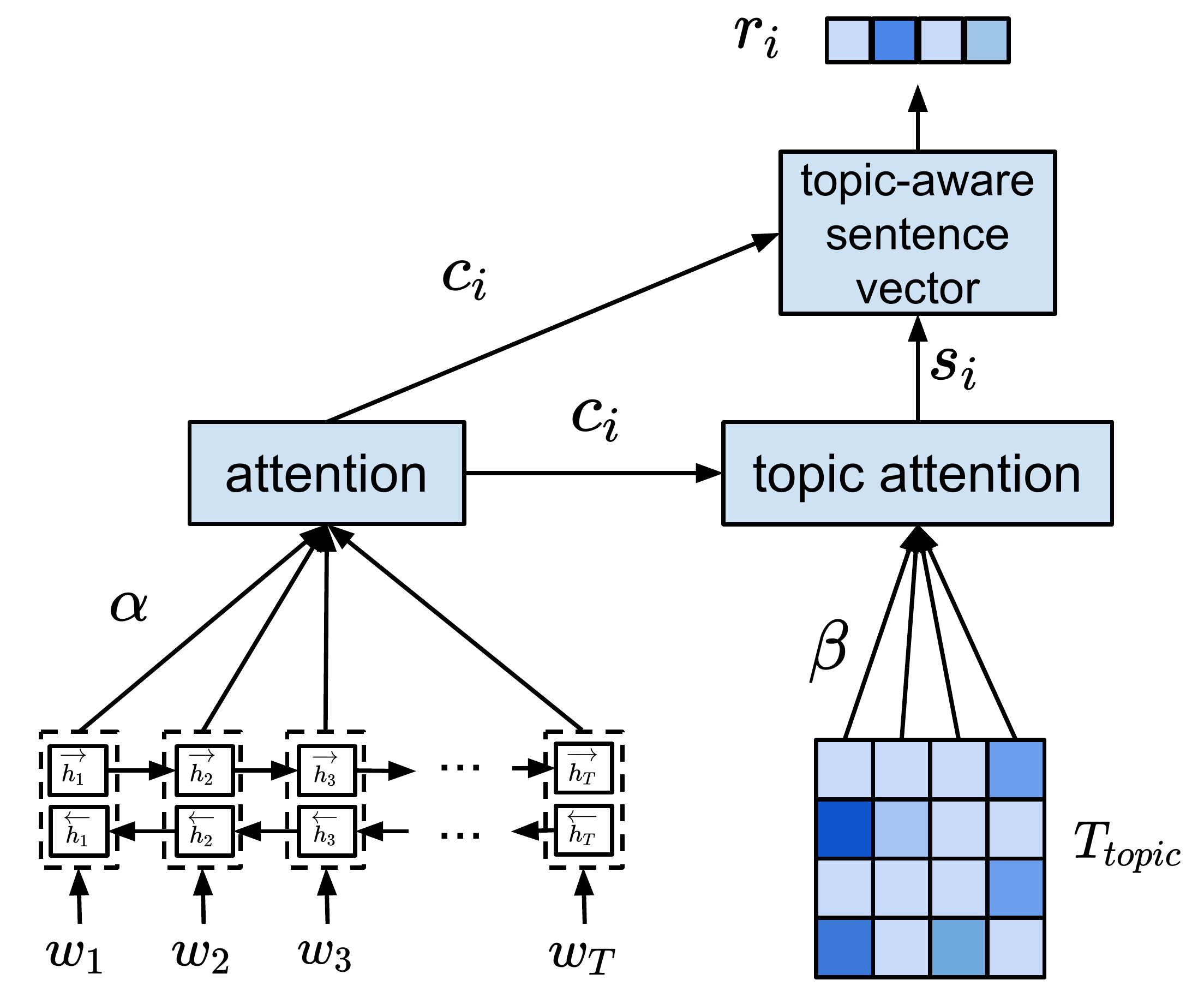}
\captionof{figure}{Topic-attention Model}
\label{model}
\end{minipage}


The last group contains the results of our proposed model with different settings. We used \textit{Topic Only} setting on top of the base model, where we only added the topic-attention layer, and all the word embeddings were from our pre-trained Doc2vec model and were fine-tuned during back propagation. We could observe that compared with the base model, we have an improvement of 7.36\% on accuracy and 9.69\% on F1 score. Another model (\textit{ELMo Only}) is to initialize the word embeddings of the base model using the pre-trained ELMo model, and here no topic information was added. Here we have higher scores than \textit{Topic Only}, and the accuracy and F1 score increased by 9.87\% and 12.26\% respectively compared with the base model. We then conducted a combination of both(\textit{ELMo+Topic}), where the word embeddings from the sentences were computed from the pre-trained ELMo model, and the topic representations were from the pre-trained Doc2vec model. We have a remarkable increase of 12.27\% on the accuracy and 14.86\% on F1 score. 
\par
To further compare our proposed topic-attention model and the base model, we report an average of area under the curve(AUC) score among all 30 terms: the base model has an average AUC of 0.7189, and our topic-attention model (\textit{ELMo+Topic}) achieves an average AUC of 0.8196. We provide a case study in Appendix \ref{app:roc}. The results show that the model can benefit from ELMo, which considers contextual features, and the topic-attention module, which brings in topic information. We can conclude that under the few-shot learning setting, our proposed model can better capture the sentence features when only limited training samples are explored in a small-scale dataset. 
\vspace{-3mm}
\section{Conclusion}
\vspace{-3mm}
In this paper, we propose a neural topic-attention model with few-shot learning for medical abbreviation disambiguation task. We also manually cleaned and collected training and testing data for this task, which we release to promote related research in NLP with clinical notes. In addition, we evaluated and compared a comprehensive set of baseline models, some of which had never been applied to the medical term abbreviation disambiguation task.  Future work would be to adapt other models like BioBERT or BERT to our proposed topic-attention model. We will also extend the method into other clinical notes tasks such as drug name recognition and ICD-9 code auto-assigning \cite{coffman2007clinical}. In addition, other LDA-based approach can be investigated.




\bibliography{neurips_bib}
\bibliographystyle{unsrt}

\clearpage

\begin{appendices}

\section{Dataset Details}
\label{appendix:graph}

\begin{figure}[th!]
\minipage{0.4\textwidth}
  \includegraphics[width=\linewidth]{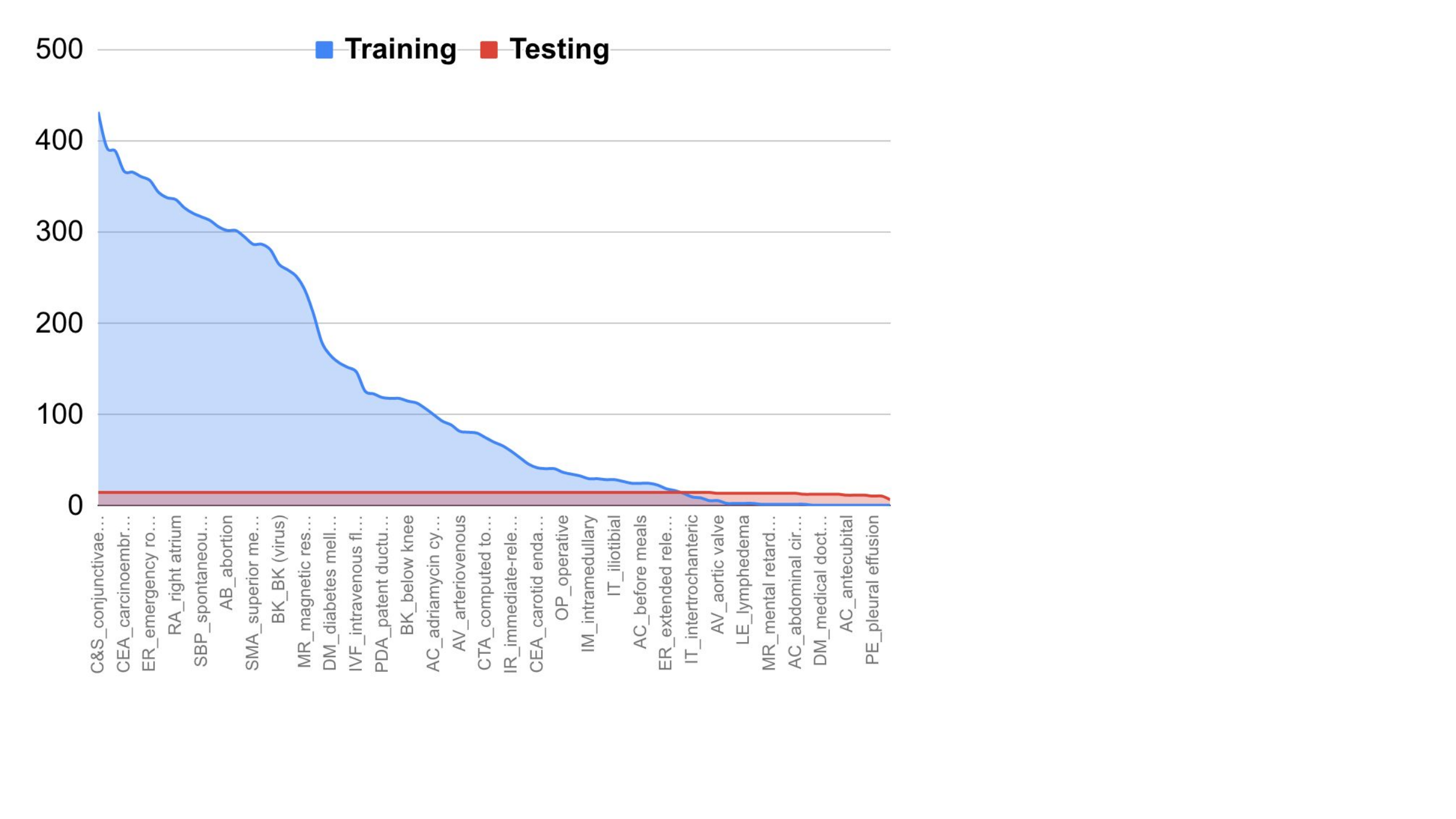}
  \caption{\small{Comparison of sample numbers. X-axis: all the 93 term-sense pairs (due to space limits, we did not list the pairs in the plot). Y-axis: how many samples in each pair.}}
  \label{fig:comp}
\endminipage\hfill
\minipage{0.25\textwidth}
  \includegraphics[width=\linewidth]{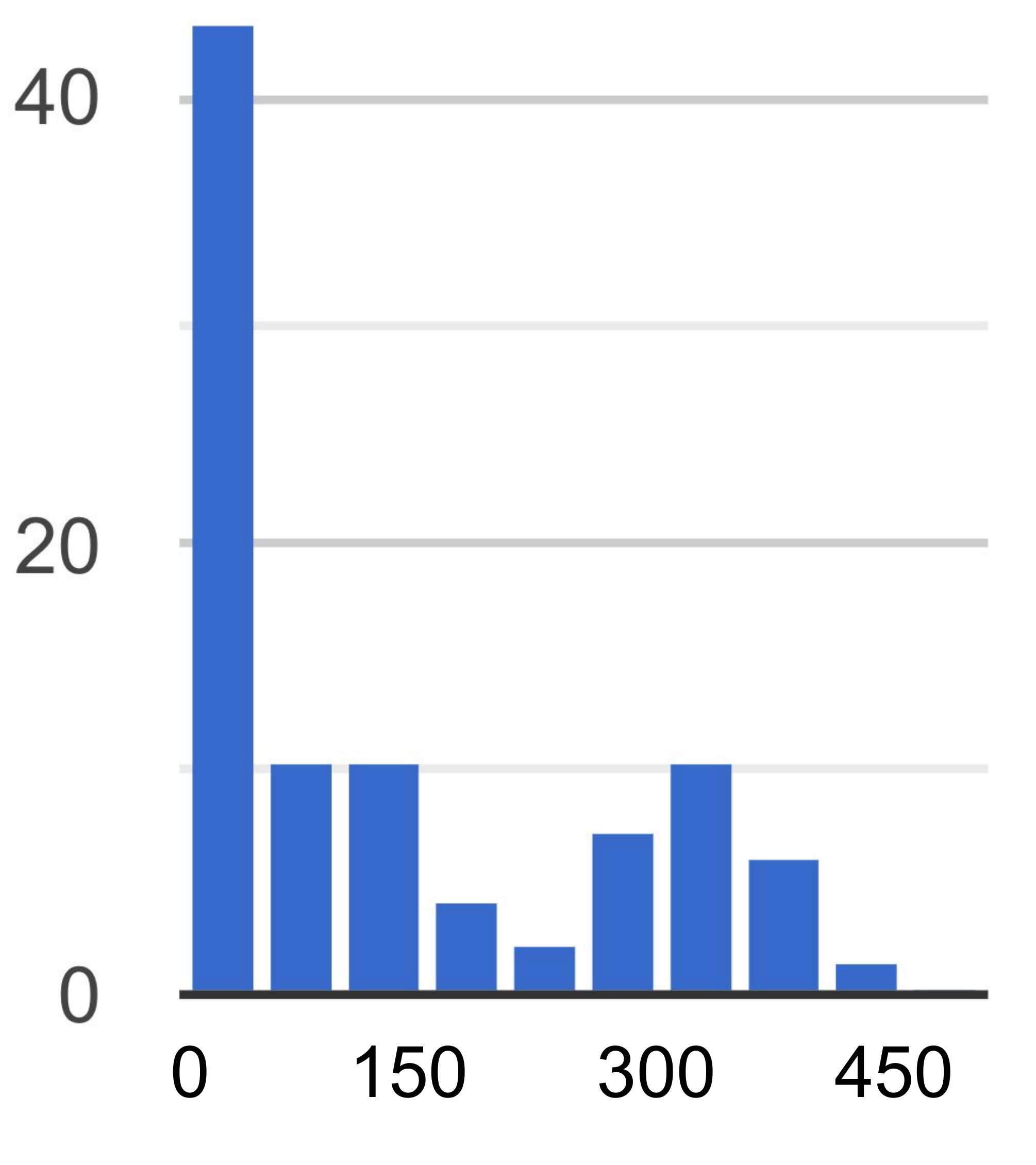}
  \caption{\small{The term-sense pair count distribution histogram among 93 pairs.}}
    \label{fig:pair}
\endminipage\hfill
\minipage{0.25\textwidth}%
  \includegraphics[width=\linewidth]{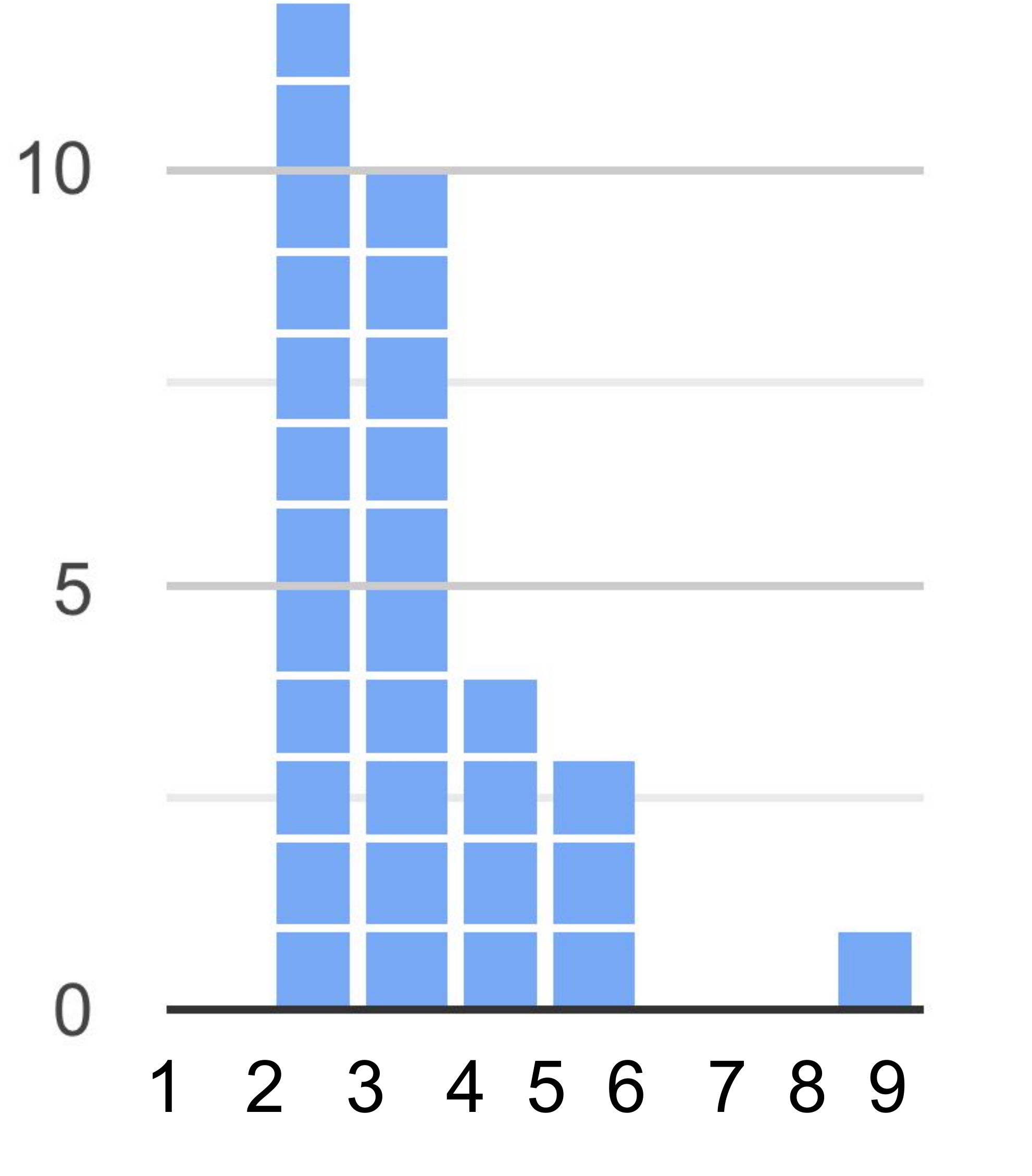}
  \caption{\small{The histogram of class numbers among all 30 terms.}}
    \label{fig:class}
\endminipage
\end{figure}

Figure \ref{fig:pair} shows the histogram for the distribution of term-sense pair sample numbers. The X-axis gives the pair sample numbers and the Y-axis shows the counts. 
For example, the first bar shows that there are 43 term-sense pairs that have the sample number in the range of 0-50. 

We also show histogram of class numbers for all terms in Figure \ref{fig:class}. The Y-axis gives the counts while the X-axis gives the number of classes. For instance, the first bar means there are 12 terms contain 2 classes. 

\section{Testing Dataset}
\label{app:testing}
Since the training dataset is unbalanced and relatively small, it is hard to split it further into training and testing. Existing work \cite{joopudi2018convolutional,xu2015clinical} conducted fold validation on the datasets and we found there are extreme rare senses for which only one or two samples exist. Besides, we believe that it is better to evaluate on a balanced dataset to determine whether it performs equally well on all classes. While most of the works deal with unbalanced training and testing that which may lead to very high accuracy if there is a dominating class in both training and testing set, the model may have a poor performance in rare testing cases because only a few samples have been seen during training. To be fair to all the classes, a good performance on these rare cases is also required otherwise it may lead to a severe situation. In this work, we are very interested to see how the model works among all senses especially the rare ones. Also, we can prevent the model from trivially predicting the dominant class and achieving high accuracy. As a result, we decided to create a dataset with the same number of samples for each case in the test dataset. 

Our testing dataset takes MIMIC-III \cite{johnson2016mimic} and PubMed\footnote{\url{https://www.ncbi.nlm.nih.gov/pubmed/}} as data sources. Here we are referring to all the notes data from MIMIC-III (NOTEEVENTS table \footnote{\url{https://mimic.physionet.org/mimictables/noteevents/}}) and \textit{Case Reports} articles from PubMed, as these contents are close to medical notes. To create the test set, we first followed the approach by \cite{joopudi2018convolutional} who applied an auto-generating method. Initially, we built a term-sense dictionary from the training dataset. Then we did matching for the sense words or phrases in the MIMIC-III notes dataset, and once there is a match, we replaced the words or phrases with the abbreviation term \footnote{For example, in our dictionary, we have the term-sense pair \textit{AC} as \textit{antecubital}. Then once the word \textit{antecubital} is found in a sentence, we replace it with \textit{AC} and let the annotators to annotate.}. 
We then asked two researchers with a medical background to check the matched samples manually with the following judgment: given this sentence and the abbreviation term and its sense, do you think the content is enough for you to guess the sense and whether this is the right sense? 
To estimate the agreement on the annotations, we selected a subset which contains 120 samples randomly and let the two annotators annotate individually. We got a Kappa score \cite{cohen1960coefficient} of 0.96, which is considered as a near perfect agreement. We then distributed the work to the two annotators, and each of them labeled a half of the dataset, which means each sample was only labeled by a single annotator. For some rare term-sense pairs, we failed to find samples from MIMIC-III. The annotators then searched these senses via PubMed data source manually, aiming to find clinical notes-like sentences. They picked good sentences from these results as testing samples where the keywords exist and the content is informative. For those senses that are extremely rare, we let the annotators create sentences in the clinical note style as testing samples according to their experiences.
Eventually, we have a balanced testing dataset, where each term-sense pair has around 15 samples for testing (on average, each pair has 14.56 samples and the median sample number is 15), and a comparison with training dataset is shown in Figure \ref{fig:comp}.  Due to the difficulty for collecting the testing dataset, we decided to only collect for a random selection of 30 terms. On average, it took few hours to generate testing samples for each abbreviation term per researcher.

\section{Case Study}
\label{app:roc}

\begin{figure}[t!]
\small
\centering
\begin{minipage}[b]{0.4\textwidth}
\includegraphics[width=\textwidth]{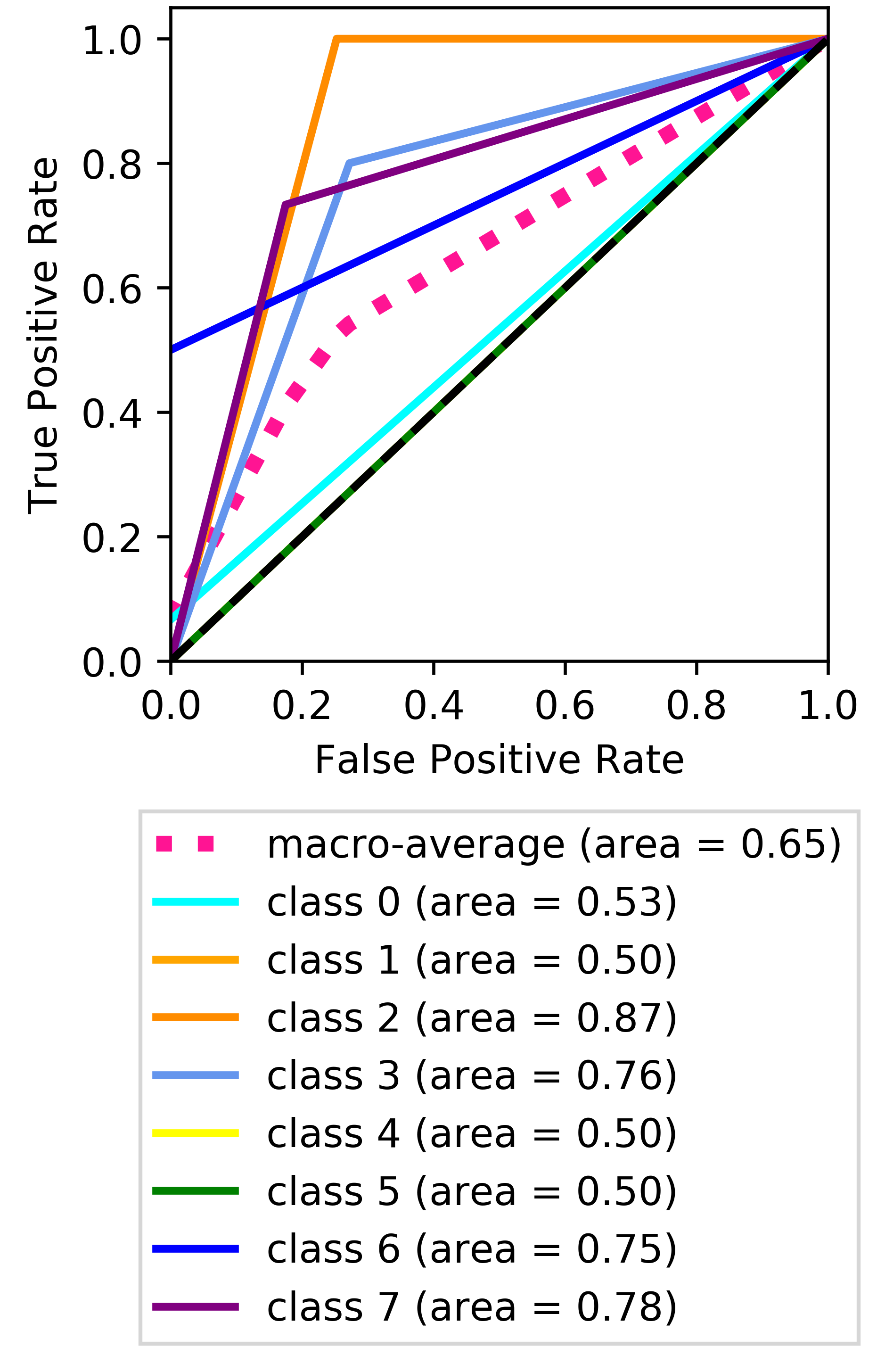}
\captionof{subfigure}{base model plot for AC}
\label{fig:ac_b}
\end{minipage}
\hfill%
\begin{minipage}[b]{0.4\textwidth}
\includegraphics[width=\textwidth]{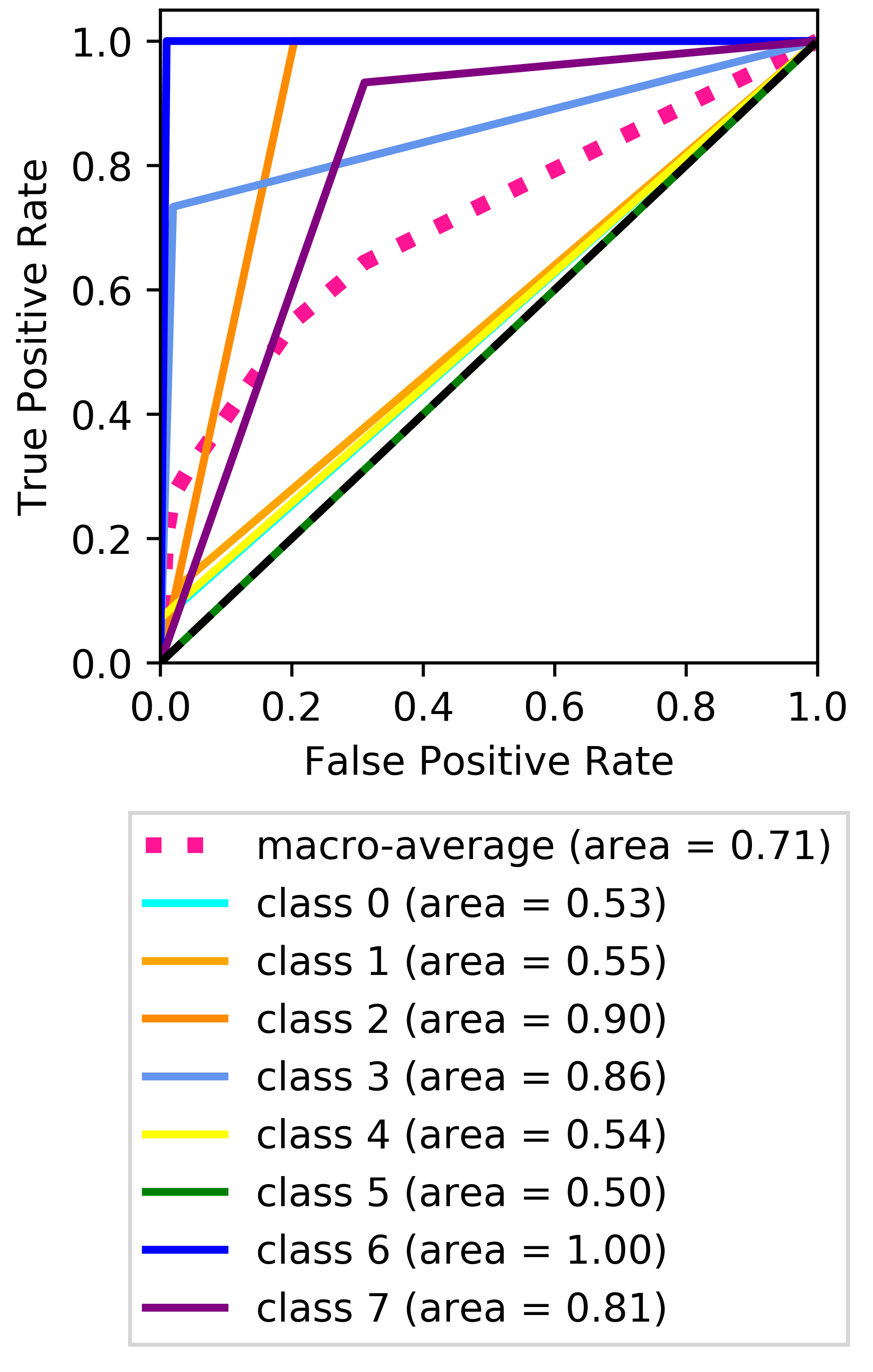}
\captionof{subfigure}{our best model plot for AC}
\label{fig:ac_a}
\end{minipage}
\hfill%
\begin{minipage}[b]{0.4\textwidth}
\includegraphics[width=\textwidth]{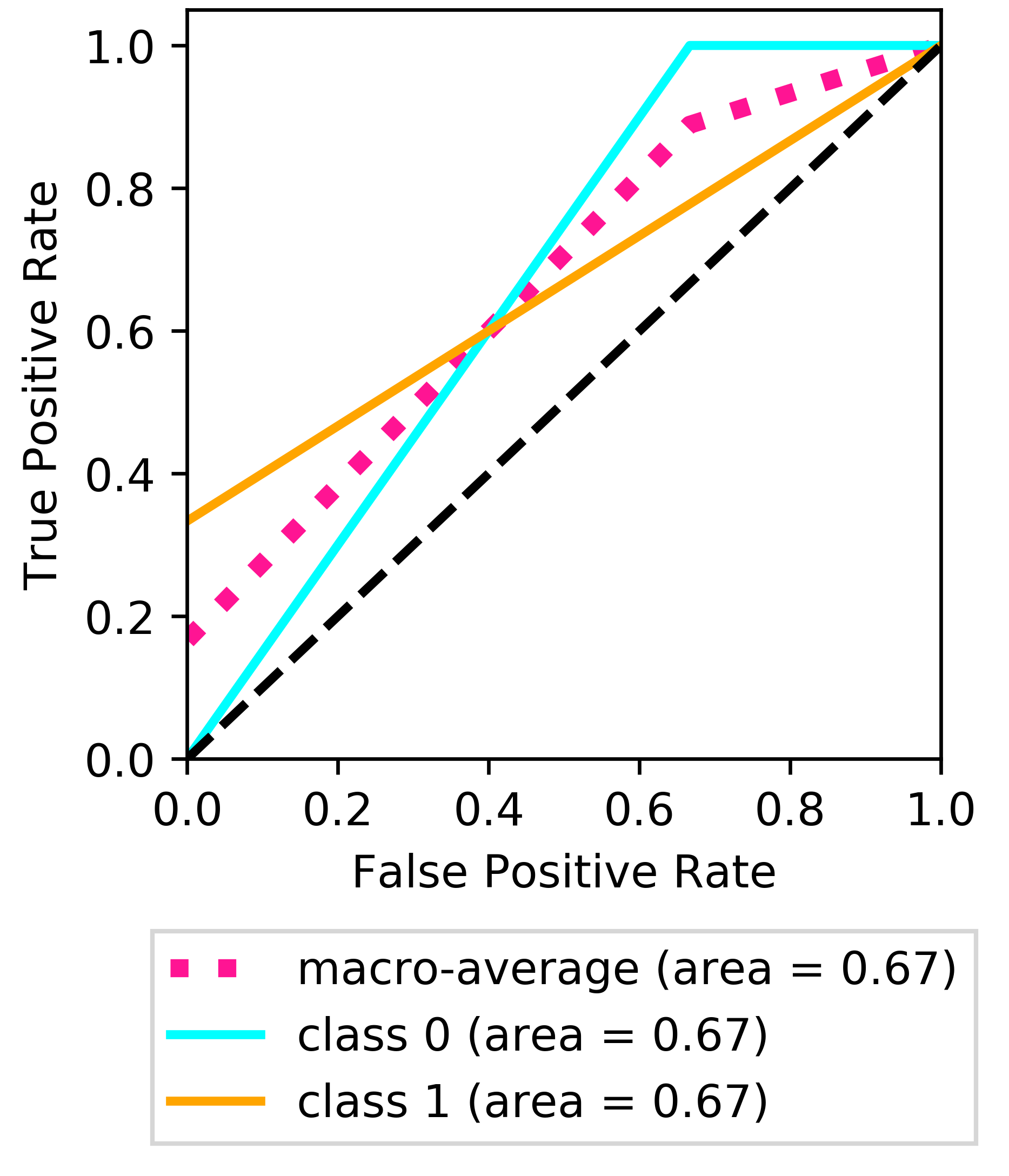}
\captionof{subfigure}{base model plot for IM}
\label{fig:im_b}
\end{minipage}
\hfill%
\begin{minipage}[b]{0.4\textwidth}
\includegraphics[width=\textwidth]{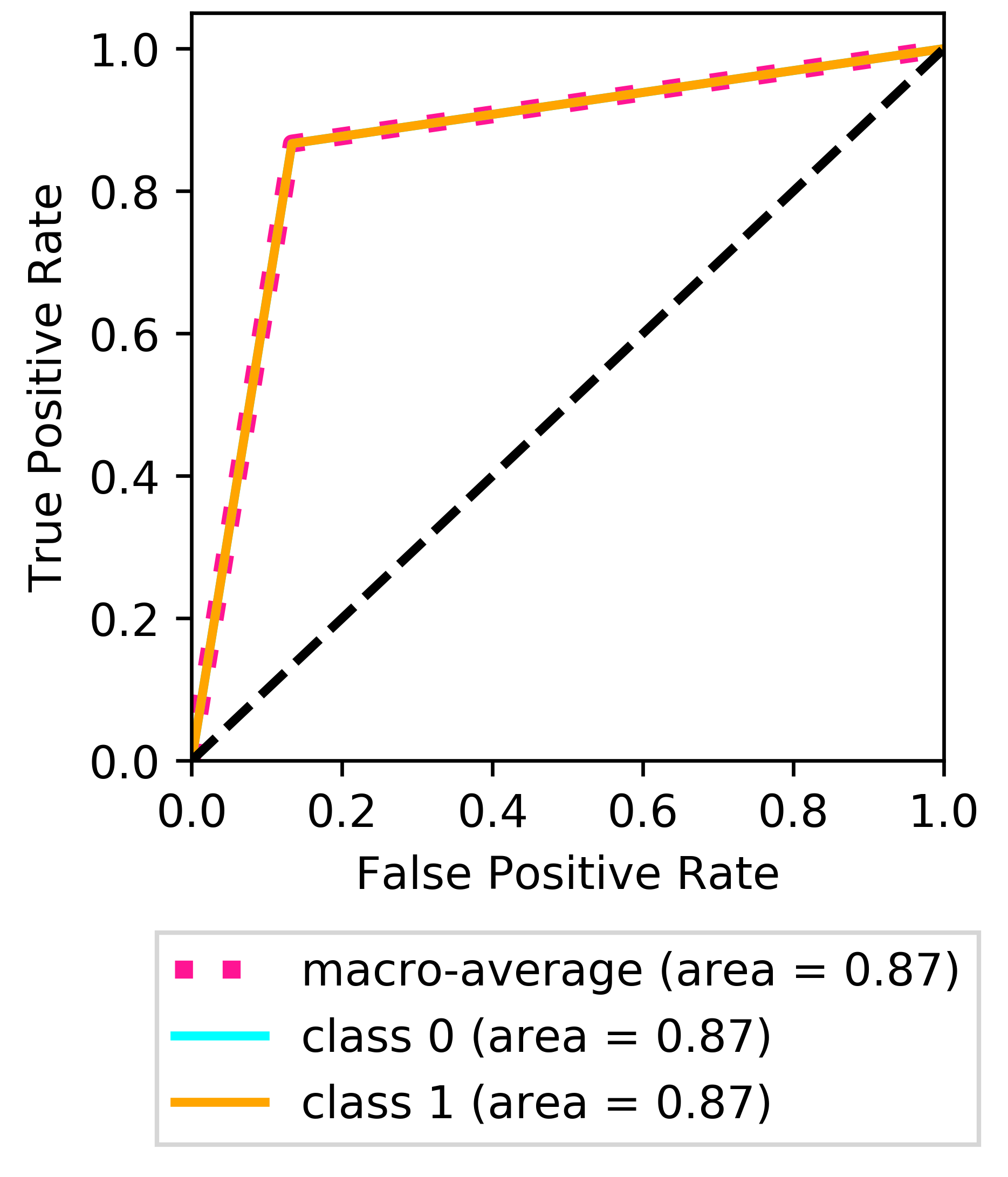}
\captionof{subfigure}{our best model plot for IM}
\label{fig:im_a}
\end{minipage}

\caption{ROC curve plots on two selected terms: AC and IM. We compare results of the base model and our best model. Here we plot the ROC curve for each class.}
\label{plot:ROC}

\end{figure}

\begin{table}
\centering
\begin{tabular}{|c|c|c|}\hline
Class ID & \# in training & \# in testing \\ \hline
\textit{AC Class} & & \\
0        & 2              & 15            \\
1        & 4              & 15            \\
2        & 158            & 15            \\
3        & 118            & 15            \\
4        & 1              & 14            \\
5        & 1              & 15            \\
6        & 9              & 14            \\
7        & 42             & 15           \\ \hline
\textit{IM Class} & & \\
0        & 38              & 15            \\
1        & 461              & 15            \\ \hline
\end{tabular}
\vspace{5mm}
\captionof{table}{Training and Testing sample distribution among classes in category \textit{AC} and \textit{IM}.}
\label{tab:ana}
\end{table}

We now select two representative terms \textit{AC}, \textit{IM} and plot their receiver operating characteristic(ROC) curves. The term has a relatively large number of classes and the second one has extremely unbalanced training samples. We show the details in Table \ref{tab:ana}.
We have 8 classes in term \textit{AC}. Figure \ref{plot:ROC}(a) illustrates the results of our best performed model and Figure \ref{plot:ROC}(b) shows the results of the base model. The accuracy and F1 score have an improvement from 0.3898 and 0.2830 to 0.4915 and 0.4059 respectively.  Regarding the rare senses (for example, class 0, 1, 4 and 6), we can observe an increase in the ROC areas. Class 6 has an obvious improvement from 0.75 to 1.00. Such improvements in the rare senses make a huge difference in the reported average accuracy and F1 score, since we have a nearly equal number of samples for each class in the testing data. Similarly, we show the plots for \textit{IM} term in Figure \ref{plot:ROC}(c) and \ref{plot:ROC}(d). \textit{IM} has only two classes, but they are very unbalanced in training set, as shown in Table \ref{tab:ana}. The accuracy and F1 scores improved from 0.6667 and 0.6250 to 0.8667 and 0.8667 respectively. We observe improvements in the ROC areas for both classes. This observation further shows that our model is more sensitive to all the class samples compared to the base model, even for the terms that have only a few samples in the training set. Again, by plotting the ROC curves and comparing AUC areas, we show that our model, which applies ELMo and topic-attention, has a better representation ability under the setting of few-shot learning.

\end{appendices}
\end{document}